\documentclass[runningheads]{llncs}
\usepackage{graphicx}
\usepackage{amsmath,amssymb} 
\usepackage{color}
\usepackage[width=122mm,left=12mm,paperwidth=146mm,height=193mm,top=12mm,paperheight=217mm]{geometry}
\usepackage{indentfirst}
\begin{document}
\pagestyle{headings}
\mainmatter
\def\ECCV18SubNumber{680}  

\title{Image Semantic Transformation: Faster, Lighter and Stronger} 


\authorrunning{Dasong Li, Jianbo Wang}

\author{Dasong Li, Jianbo Wang}
\institute{Shanghai Jiao Tong University, ZheJiang University}

\maketitle

\begin{abstract}
We propose Image-Semantic-Transformation-Reconstruction-Circle(ISTRC) model, a novel and powerful method using facenet's Euclidean latent space to understand the images. As the name suggests, ISTRC construct the circle, able to perfectly reconstruct images. One powerful Euclidean latent space embedded in ISTRC is FaceNet's last layer with the power of distinguishing and understanding images.
Our model will reconstruct the images and manipulate Euclidean latent vectors to achieve semantic transformations and semantic images arthimetic calculations.
In this paper, we show that ISTRC performs 10 high-level semantic transformations like "Male and female","add smile","open mouth", "deduct beard or add mustache", "bigger/smaller nose", "make older and younger", "bigger lips", "bigger eyes", "bigger/smaller mouths" and "more attractive". It just takes 3 hours(GTX 1080) to train the models of 
10 semantic transformations.


\end{abstract}

\section{Introduction}
Ideally, we want to derive a Euclidean space as the powerful latent space to understand images. Because the distance between two images in this latent space could be calculated, we could implement classification, verification and other tasks in this latent space. 

But the spaces in GANs are designed for the tasks of generating pictures(for example uniformed distribution latent space in DCGAN\cite{DCGAN}, gaussian distribution latent space in Progressive Growing GAN\cite{PGGAN}). The latent space is not interpretable because the latent space is in manifold structure\cite{manifold}. There is another type of GANs, called auto-encoder GANs. In this kind of GANs, Discriminator is able to reconstruct the images. 

The FaceNet\cite{FaceNet} provides a perfect model to encode the images to their latent vectors for accurate image classification and recognization. We take the output of the FaceNet as the Euclidean latent space. Many magnficient functions such as understanding images and semantic transformations are explored in this Euclidean latent space.

Image-Semantic-Transformation-Reconstruction-Circle(ISTRC) make efforts on combining the advantage of Euclidean latent space and auto-encoder GANs. One mapping are used to connect the FaceNet and BEGAN to manipulate the Euclidean latent space for semantic editing and reconstruct the image with BEGAN. At first, FaceNet takes one image as input and output one latent vector in Euclidean latent space. Secondly, we could operate some semantic transformation in the Euclidean latent space. At last, the latent vectors generated by the transformations are sent to Decoder of Discriminator of BEGAN to reconstruct the images.

Our key insight is to use gradient descent to increase the probability that the latent vector could be judged to have specific character. In this process, we view the images generated by these gradually changing latents. It shows these images also indeed closing to the target visual concept.
In this paper, 10 semantic transformation models are trained in 3 hours. All the transformations have shown perfect performances.




\begin{figure}
\centering
\includegraphics[height=3cm]{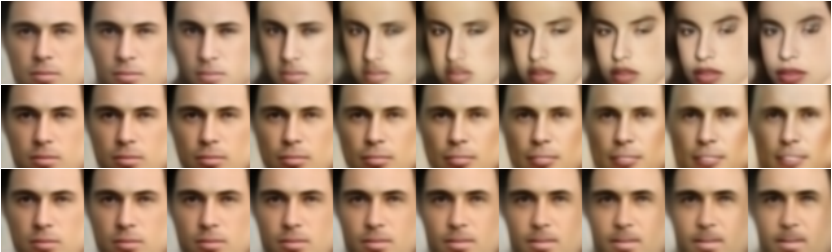}
\caption{They are the semantic transformations of Chris Evans. From top to bottom, they are adding makeup, bigger lips and adding mustache. Leftmost images are input images}
\label{fig:example}
\end{figure}

\section{Related work}

A large amount of effort has been devoted to applying arithmetic in latent space to change visual concept for the produced image. Here is a list of related work. 

\subsection{Understanding images on latent space in GANs}

Nowadays, generative models for image data are highly popular, including the Variational Autoencoder\cite{VAE} and GANs\cite{GAN}. Among them, GANs is a common way for generating convincing images. DCGAN\cite{DCGAN} changes the structure of vanilla GANs, which uses convolutional structure instead of full connected networks. The fascinating experience is that they try to do vector arithmetic. Like "smiling woman - neutral woman + neutral man" to get a "smiling man". BEGAN \cite{BEGAN} set the discriminator in auto-encoder format, which also improves the stability in training process. 

The latent spaces in DCGAN\cite{DCGAN} and BEGAN\cite{BEGAN} are in uniform distribution in the range of [-1,1]. They both have operated linear interpolation between some random points in their latent spaces. 
\begin{figure}
\centering
\includegraphics[height=5cm]{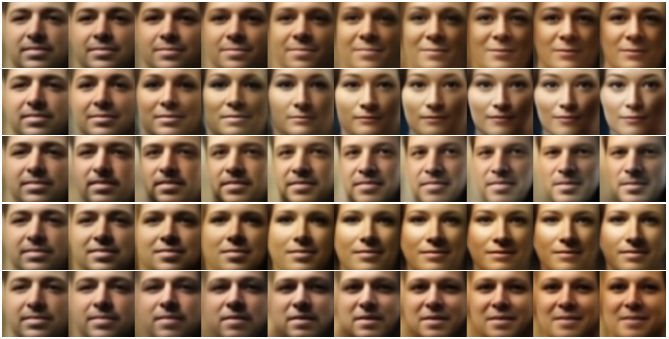}
\caption{They are five semantic transformations of the same person. From top to bottom, they are less beard, adding makeup, adding Mustache, more attractive and adding smile. Leftmost images are input images.}
\label{fig:example}
\end{figure}
Many GANs\cite{InfoGAN} are trying to understand the latent space in vector, which makes sense at a high cost of time. But if the network's structure is changed, the latent space in GANs will also be changed. 
We introduce one perfect Euclidean latent space to entitle GANs the power of understanding images.

\subsection{Semantic transformations}

The most similar work to this paper is Deep Feature Interpolation for image content changes\cite{DeepFeature}. In this paper, their model has successfully changed the character of faces. Features extracted by vgg\cite{VGG} are used to operate the interpolation to accomplish the semantic interpolation. Since there is no closed-form inverse function for Vgg's mapping, the reverse mapping of gradient descent costs much time. Our Model is more lighter and faster than it.

\subsection{FaceNet}
FaceNet\cite{FaceNet} can directly learn the mapping from images to their latent vector in Euclidean space end-to-end. It is used for face verification since we could compare the similarity between two face images by their Euclidean distances in this Euclidean space. 
In this paper, we directly use the pre-trained FaceNet model for cropped CelebA images.

\begin{figure}
\centering
\includegraphics[height=6cm]{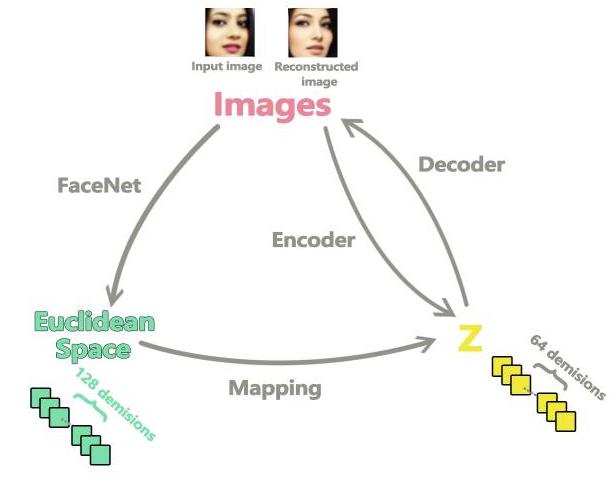}
\caption{Structure of Image-Semantic-Transformation-Reconstruction-Circle.}
\label{fig:example}
\end{figure}
\section{Image-Semantic-Transformation-Reconstruction-Circle}
We propose Image-Semantic-Transformation-Reconstruction-Circle to deeply understand images. All the circle consists of four main components -- FaceNet, Euclidean latent space, Discriminator of auto-encoder GANs and the mapping from Euclidean latent space to the latent space of GANs.
\subsection{FaceNet}
In FaceNet, the triplet selection is applied as the core method to train all the faces into one Euclidean space. All the process is in the supervised learning.
The output vector of FaceNet is embedded to live on a 128-dimensional hypersphere where all the vectors $Z$ have $||Z||_2 = 1$. This hypersphere is the Euclidean latent space. Because the distance of two vectors in this space represent the similarity of two images, we could find more brilliant functions there, such as learning other features and semantic transformations.

\subsection{BEGAN}

BEGAN is a well-known auto-encoder GAN to generate the faces images. The discriminator in BEGAN could encode the input image onto latent space and reconstruct the images. The latent space is in a Manifold structure\cite{mani}, which means the distance between two latent vector could not be taken as the actual intrinsic distance. And Our model only uses the part to help Z to images reconstructed.
\begin{figure}
\centering
\includegraphics[height=4cm]{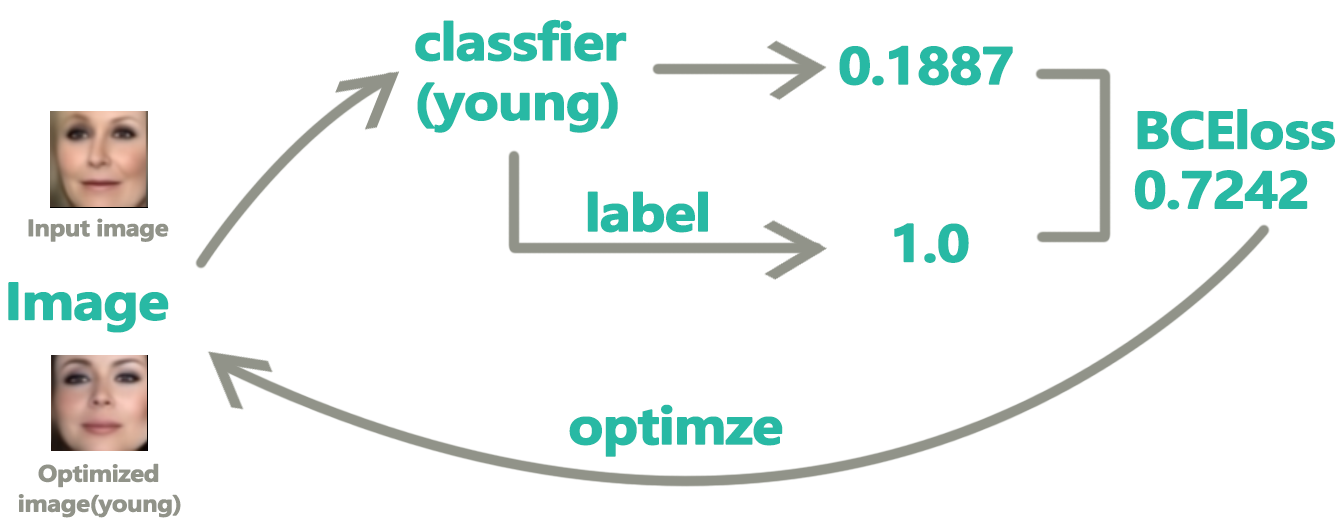}
\caption{The process of optimize the Euclidean latent vector to target.}
\label{fig:example}
\end{figure}
\subsection{The Mapping}
Now that Euclidean latent space has great power of understanding images and BEGAN has the great power to reconstruct images, the mapping serves as the bridge to connect Euclidean latent space and BEGAN to achieve many fantastic functions in ISTRC. We have tried to learn the mapping from Euclidean latent space to the initialized latent space of GANs. But the loss is so high that we have to give it up. And the mapping from Euclidean space to latent space of Discriminator in BEGAN is pretty easy to train. We define the set of Euclidean latent vectors as Z and the set of latent space of Discriminator in BEGAN as $Z_2$. The mapping is $F(Z) \rightarrow Z_2$ and the cost function in training is the Mean Square Error (MSE) with regularization. In practice, we find the structure with fully connected layers are more useful than convolutional neural network in learning the mapping between $Z$ and $Z_2$. The mapping connect two component and form the circle. The structure of the ISTRC is shown in Fig 3.
\begin{figure}
\centering
\includegraphics[height=3cm]{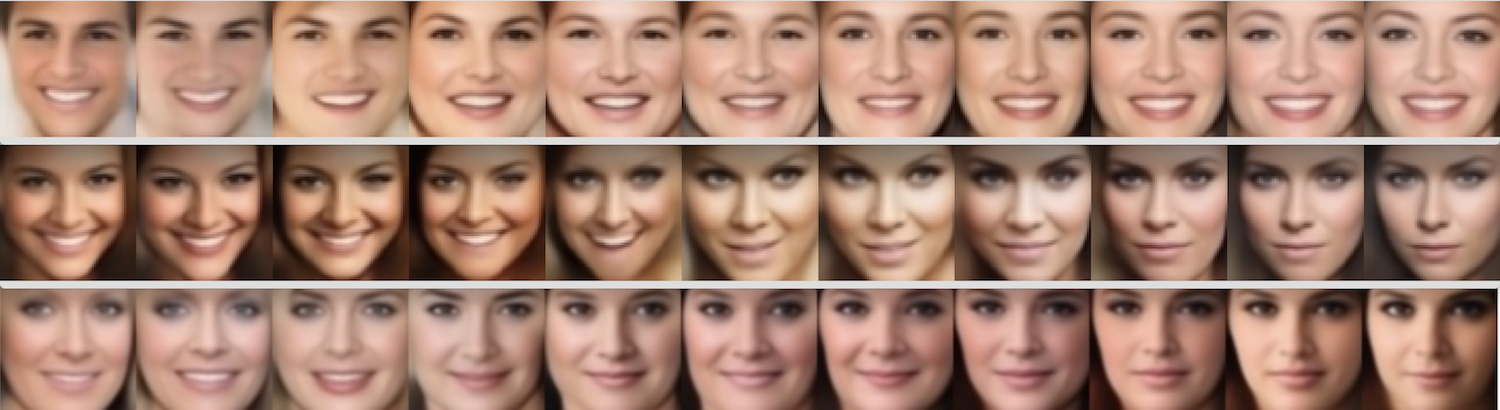}
\caption{Interpolation results in Euclidean latent space.}
\label{fig:example}
\end{figure}

\subsection{Euclidean Latent Space}
At first, this Euclidean latent space is trained for image classification and verification and recognization. We explore more functions in this space. This space should support arthimetic calculation and average images. We will talk about this in next section.
what's more, classifiers of many images features could be easily trained in this space\cite{FicialClass}. After getting the classifiers, we could let the latent vector z move through the gradient to change feature. The loss function is 
\begin{equation}
loss(z,y) = -[ylog z + (1-y)log(1-z)]
\end{equation}
$y$ could be 0 or 1 which represents two opposite directions in semantic transformation. The learning rate is dynamically adjusted to ensure that vector moves the same distance is constant every iteration. Also the vector will be normalized ($||z||_2 = 1$) every iteration. I think the neural network could automatically help us to find which dimensions are efficient\cite{hinton2006reducing} for some assigned features such as big nose.
For instance, a classifier of judging whether a face is smiling will give the latent vector of a neutral face low score. The latent vector will go through the gradient to achieve as high score as possible. Iterations will witness the gradually changing from neural face to smiling face. With this idea, 10 classifiers are trained and many semantic transformations are easily accomplished. Although this Euclidean latent space is very powerful, it must be highlighted that without the Circle to reconstruct images, all the functions in this space could never be demonstrated and proven.

\begin{figure}
\centering
\includegraphics[height=6cm]{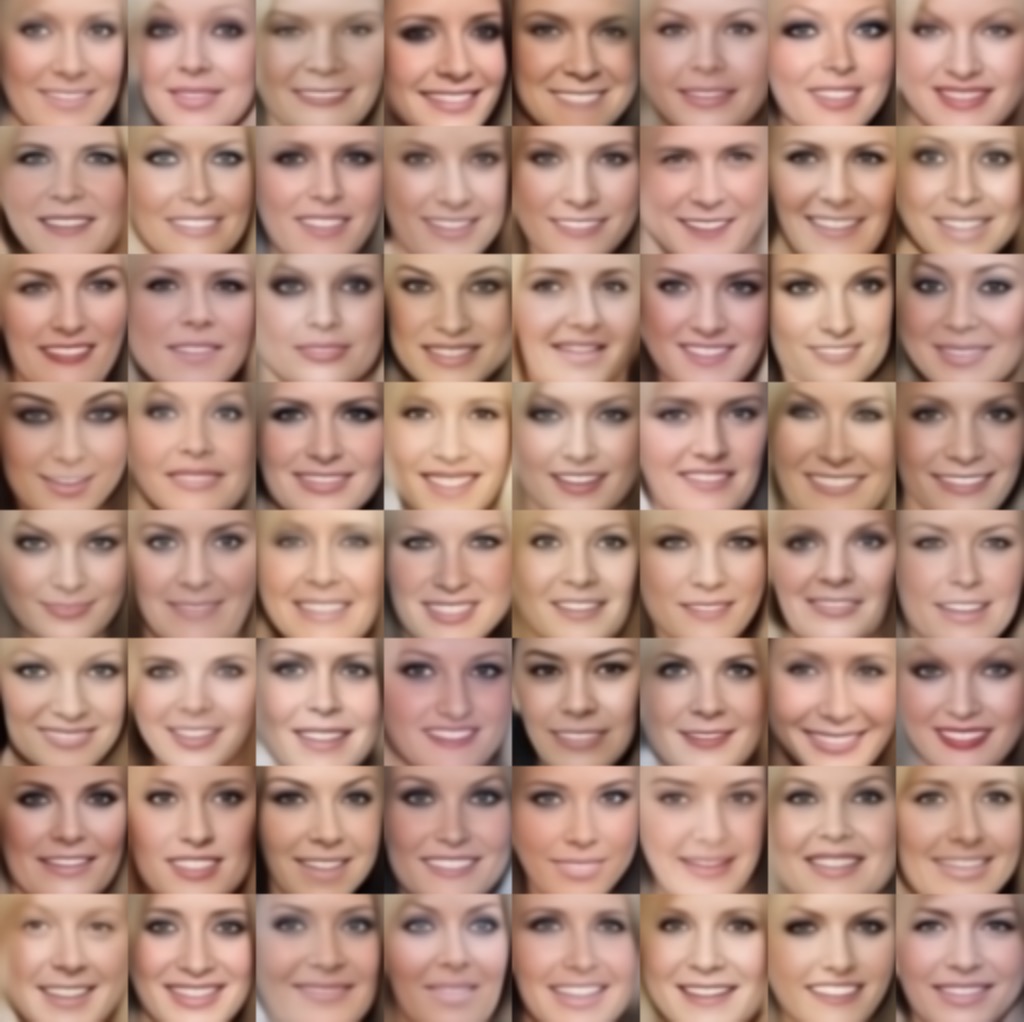}
\caption{The image (1,1) add 63 small random noise vectors generate slightly different faces.}
\label{fig:example}
\end{figure}
\section{Evaluating Euclidean latent space}
In this space, the Euclidean distance represent the similarity of images. So we can achieve many recognization tasks such as classification, verfication and etc. A little different to the FaceNet, in our space, the distance is calculated by the high dimensional spherical distance. In the cropped CelebA dataset\cite{CelebA}, our method performs a little greater than the original facenet. The reason why we need to use this high dimensional spherical distance is that in the distance should be precise to finish the following experiments precisely.
\subsection{Proof of space's smooth and connectivity}
In order to prove our Euclidean space smooth and be connected with each other, we use the images interpolation\cite{interpolation} to illustrate this point.
In our Euclidean latent space, spherical linear interpolation(slerp)\cite{Slerp} and linear interpolation methods are token and they achieve the same results.

Apart from the interpolation, some different small random noises are added to one Euclidean latent vector. ISTRC will reconstruct them to the images(Figure 6). All the images are slightly different, which demonstrate this space is great. 
\begin{figure}
\centering
\centering
	\begin{tabular}{cccc}
		\includegraphics[width=0.2\linewidth]{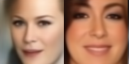}& 
		\includegraphics[width=0.1\linewidth]{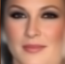}&
		\includegraphics[width=0.4\linewidth]{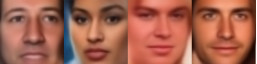}&
		\includegraphics[width=0.1\linewidth]{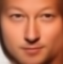} \\ 
		(a) & (b) & (c) & (d)\\
	\end{tabular}
\caption{Using sphere to calculate the average face.image (b) is the average of two images in (a). image (d) is the average of four images in (c)}
\label{fig:example}
\end{figure}

\subsection{Average image}
At first, we all know that latent space in GANs don't have the power to support the complex calculations. To find the reason, latent vector is initialized to be in uniform distribution or guassian distribution. When it comes to average face of many peoples$\{z_1,z_2,...,z_n\}$, the average latent vector $avg_z$ will be near to the gaussian distribution with very small standard deviation.
In this condition, the average face will be to the same woman when the number of z is higher than 60.
But in our Euclidean latent space, we first map all the latent set to $\{z_1',z_2',...,z_n'\}$ and then calculate the average vector $z_{avg}^{'}$. The specific formula of average vector is the spherical linear interpolation $slerp$. The formula is introduced by (sheomake\cite{Slerp}):
\begin{equation}
slerp(q_1,q_2;\mu) = \frac{sin(1-\mu)\theta}{sin\theta}q_1 + \frac{sin\mu\theta}{sin\theta}q_2
\end{equation}
The average of two Euclidean latent vector $q_1$ and $q_2$ is equal to $slerp(q1,q2;0.5)$. 
The $z_{avg}^{'}$ is mapping to the latent space of BEGAN and then generate the average face. In this way, the average face will never not be the same person. We find, the more faces are calculated, the closer to zero results of linear interpolation are. But the spherical linear interpolation $slerp$ always show dominant performance.
\begin{figure}
\centering
\centering
	\begin{tabular}{cc}
		\includegraphics[width=0.41\linewidth]{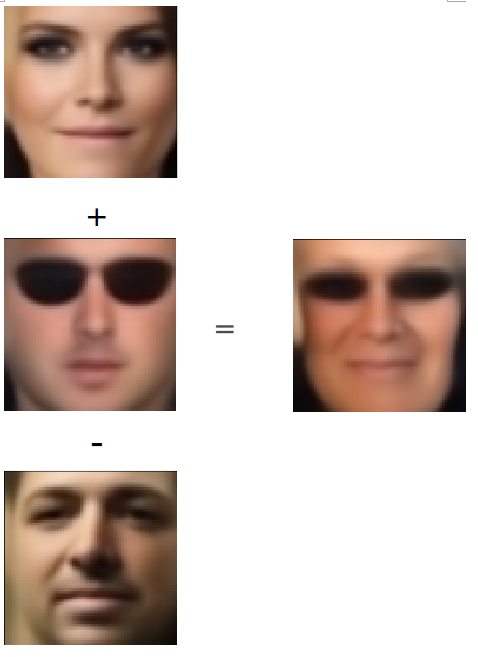}& 
		\includegraphics[width=0.4\linewidth]{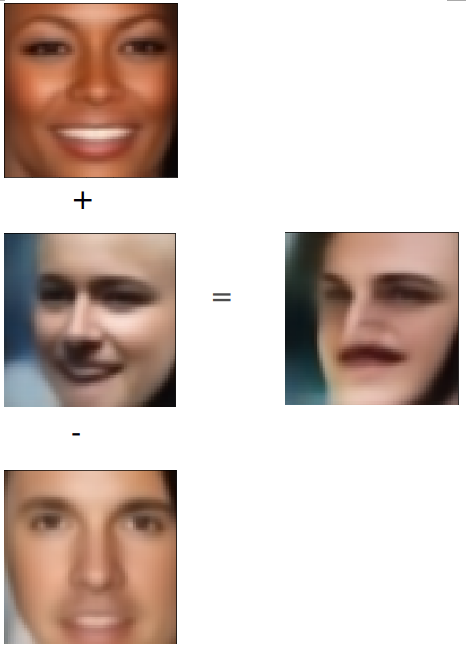} \\ 
		(a) & (b) \\
	\end{tabular}
\caption{They are two set of arithmetic results. The result faces make sense semantically.}
\label{fig:example}
\end{figure}
\subsection{Arithimetic calculation}
Pretty similiar to the above idea, mapping the latent vectors to our Euclidean latent space and calculating this equation there can help us find the accurate vector in the interpretable space and generate the right images. The Spherical linear interpolation should be used. When the equation is easy, linear calculation shows the same performance.



\section{Experiment Results}

In our paper, we evaluate ISTRC on CelebA cropped dataset. Faces images are cropped and aligned to 64*64 images only faces included. And the number of the cropped faces is about 170,000.  
\subsection{FaceNet and BEGAN}
We use the FaceNet pre-trained model which is trained in for cropped face images. The FaceNet has a great performance in face classification and recognization. As another crucial component, BEGAN is trained on this cropped faces. For the particular loss function in BEGAN helps the training pretty stable in generating human being's faces. About 10 epochs, BEGAN has generated pretty clear images. Although there are some spots in faces generated by BEGAN, the discriminator show its function in filtering the spots of input images and reconstructing the perfect images. Because there are merely a few old faces, our BEGAN has some problems in generating old faces. But this little fault will not cause too much interruption to our whole model. We also could distinguish the faces' becoming older.
\begin{figure}
\centering
\centering
	\begin{tabular}{ccc}
		\includegraphics[width=0.49\linewidth]{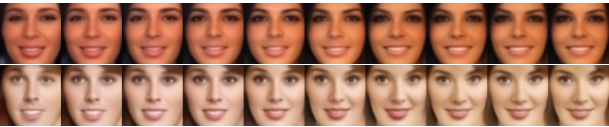}  & 
		\includegraphics[width=0.49\linewidth]{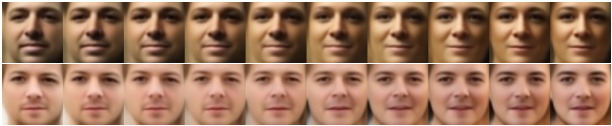} \\ 
		more attractive & less beard \\
        \includegraphics[width=0.49\linewidth]{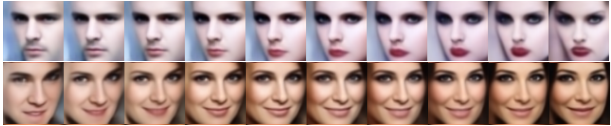}  & 
		\includegraphics[width=0.49\linewidth]{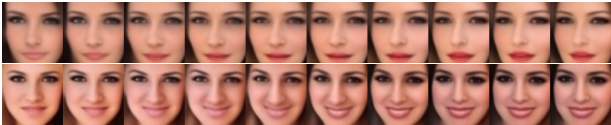} \\ 
		female & adding makeup \\
        \includegraphics[width=0.49\linewidth]{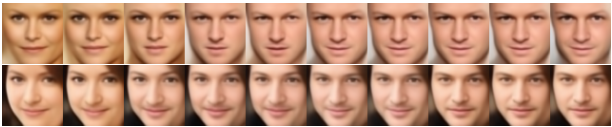}  & 
		\includegraphics[width=0.49\linewidth]{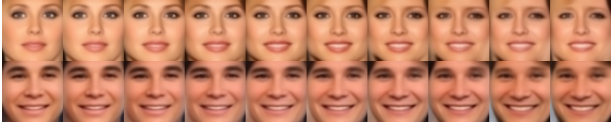} \\ 
		male & slightly open mouth \\
	\end{tabular}
\caption{Semantic transformations. Leftmost images are input images.}
\label{fig:example}
\end{figure}
\subsection{Mapping from FaceNet to BEGAN}
As is mentioned above, the output of FaceNet is in an Euclidean 128-dimensional space. The mapping is from this Euclidean high dimensiona space to the latent space of Discriminator in auto-encoder GANs. The loss of the mapping is MSE.

At first, we adapted Convolutional Neural Network to learning the mapping but the loss never descends. Then fully connected layers are tried. Five fully connected layers with Batch Normalization and tanh activation work pretty great on learning this mapping. Maybe for the mapping from vector to vector, structure with fully connected layer are predominant. Since then, the circle is connected. Inputing some images, this model will reconstruct them again. The results of images reconstruction are shown in Figure 3.

\subsection{Training classifiers in Euclidean latent space}
In the process of reconstruction, we derive the powerful latent space where the vectors represent the features. We select 10 semantic features in human faces. One classifier is trained for each feature. We know that we could train a classifier on pixel images, using convolution neural network. In our Euclidean latent space, the fully connected layers also work very well. There are about four to seven fully connected layers for different features. The more complicated one semantic feature is, the more layers are need.  

\subsection{Semantic Transformation}
With the help of ISTRC and 10 classifiers, given one image, its Euclidean latent vector use the gradient descent to let the classifier to judge the features changing. In the process of vector's gradient descent, the changing vectors are collected and reconstruct the faces. 
All the transformations work fairly magnificently and the results are shown in picture in Fig 9, Fig 10.  
\begin{figure}
\centering
\centering
	\begin{tabular}{ccc}
        \includegraphics[width=0.49\linewidth]{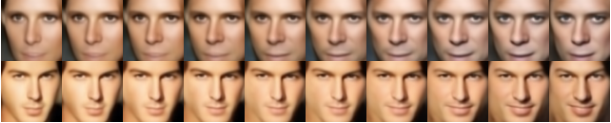}  & 
		\includegraphics[width=0.49\linewidth]{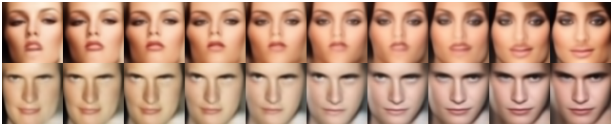} \\ 
		bigger nose & bigger eyes \\
		\includegraphics[width=0.49\linewidth]{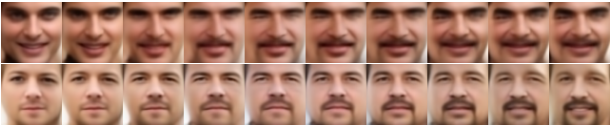}  & 
		\includegraphics[width=0.49\linewidth]{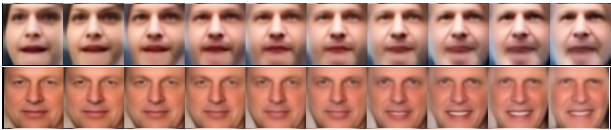} \\ 
		more mustache & older \\
        \includegraphics[width=0.49\linewidth]{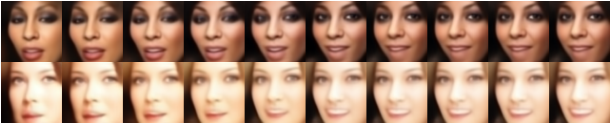}  & 
		\includegraphics[width=0.49\linewidth]{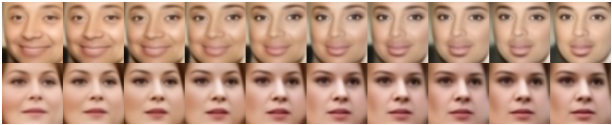} \\ 
		adding smile & younger \\
	\end{tabular}
\caption{Semantic transformations. Leftmost images are input images.}
\label{fig:example}
\end{figure}
\subsection{Empirical Thinking}
Our model is not that perfect. There are some problems for some features.
In practice, some crucial features such as makeup and beard can change men to women and change women to men(Fig 11). Based on our model, they make sense because the classifier is trained by all the people and men tend to have beard and woman tend to have makeup. We did not separate the images by the gender. The result has shown that almost all the semantic transformations make sense and show great performance. For the limited pages, More transformations are placed on Appendix. 

\section{Discussion}
\subsection{Time and space complexity}
The cost of time and space in our method could be the most lean. With the pretrained FaceNet model and one sophisticated BEGAN, we spent 2 hours and 2 GB GPU training a neural network to learn the mapping from Euclidean latent space to latent space of GANs. All the process in the circle mentioned above is linear and pretty time-efficient. Using this complete models, about 100 64 * 64 images take 1s to reconstruct. 
It cost us 3 hours in GPU to train 10 features and run the 10 classifier of 10 features at the same time.(Because classifier is operating the classification in Euclidean latent vectors, one training process only needs about 0.7 G GPU and one gtx 1080 could hold 10 processes simultaneously.) 
Given all the trained model, One image just takes 25s to iterate 500 times to reach the target semantic transformation and generate 10 transforming images in 16G CPU and without GPU.

\subsection{Our Model's simplicity}
Deep convolution neural network is only used in the FaceNet and BEGAN. The mapping from the Euclidean latent space to latent space of BEGAN is just fully connected layers. Also all the classifiers' structures are simple fully connected layers and some activation layers. With input one face image, our model could get the images' Euclidean latent vector and translate the vector back to the face image. To sum up, all the processes including the FaceNet-GAN circle and training classifier are pretty time-efficient. 
\begin{figure}
\centering
\centering
	\begin{tabular}{cc}
        \includegraphics[width=0.49\linewidth]{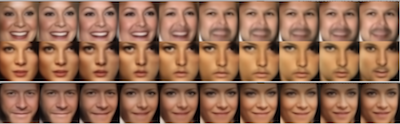}  & 
		\includegraphics[width=0.49\linewidth]{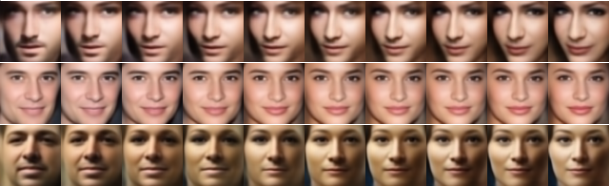} \\ 
		adding and deducting beard & adding makeup \\
	\end{tabular}
\caption{Leftmost images are input images. Sex faults in semantic transformations. But it also makes sense. }
\label{fig:example}
\end{figure}
\subsection{Beyond Face dataset}
The Euclidean latent space has a good understanding of images. The triplet selection\cite{FaceNet} could also be used in many different dataset to derive their own Euclidean latent spaces. Our model will work pretty well on that kind of dataset.
 
\section{Conclusion and Future Work}
We have introduced and demonstrated 's great performance of understanding the images and fast semantic transformations. ISTRC could understanding the images in Euclidean latent space, manipulate Euclidean latent vector to execute semantic transformation and reconstruct the images from vectors. In the future, we will improve ISTRC to process and reconstruct the higher-resolution images. Besides these 10 semantic transformation, ISTRC is believed to have the potential to achieve all 
semantic transformations of high-resolution images.


\bibliographystyle{splncs}
\bibliography{egbib}
\newpage
\begin{appendix}
\section{Appendix For Semantic Transformations}
\begin{figure}
\centering
\centering
	\begin{tabular}{c}
        \includegraphics[width=0.8\linewidth]{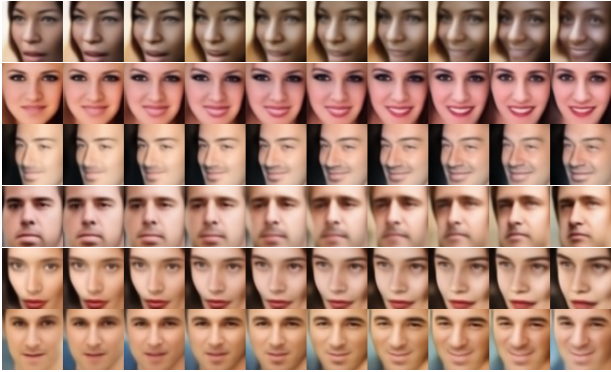}  \\ 
		bigger nose  \\
        \includegraphics[width=0.8\linewidth]{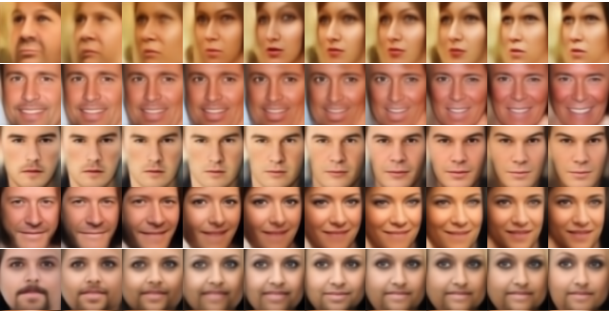}  \\ 
		less beard  \\
	\end{tabular}
\caption{Semantic transformations in supplementary material. Leftmost images are input images.}
\label{fig:example}
\end{figure}

\begin{figure}
\centering
\centering
	\begin{tabular}{c}
        \includegraphics[width=0.8\linewidth]{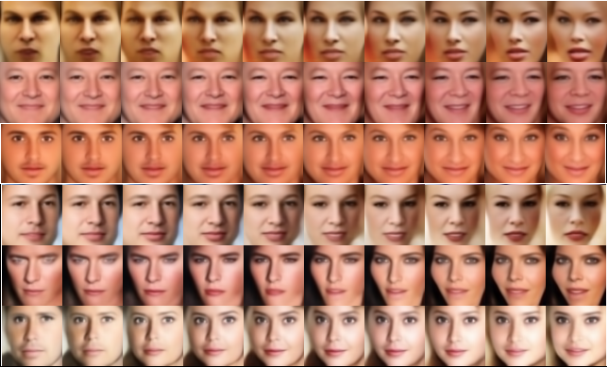}  \\ 
		female  \\
        \includegraphics[width=0.8\linewidth]{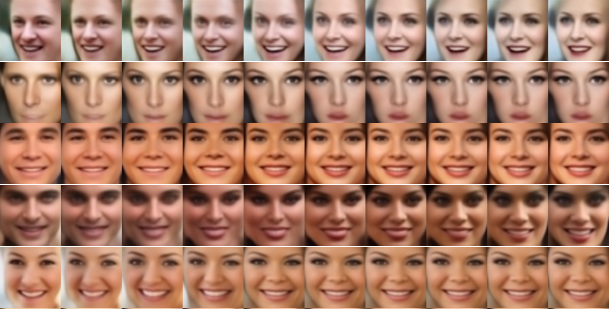}  \\ 
		adding makeup  \\
        \includegraphics[width=0.8\linewidth]{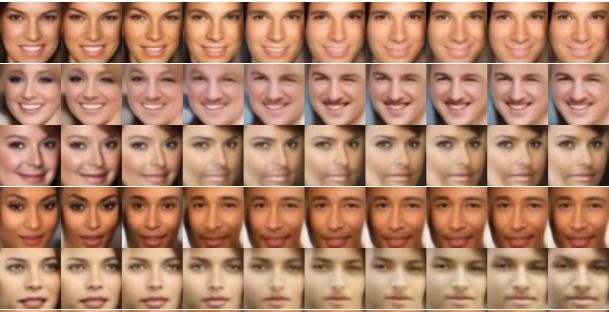}  \\ 
		male \\
	\end{tabular}
\caption{Semantic transformations in supplementary material. Leftmost images are input images.}
\label{fig:example}
\end{figure}


\end{appendix} 
\end{document}